\documentclass[letterpaper]{article} 
\usepackage{aaai2026}  
\usepackage{times}  
\usepackage{helvet}  
\usepackage{courier}  
\usepackage[hyphens]{url}  
\usepackage{graphicx} 
\usepackage{natbib}  
\usepackage{caption} 
\usepackage{algorithm}
\usepackage{algorithmic}

\usepackage{newfloat}
\usepackage{listings}
\usepackage{amsmath}
\usepackage{amsthm}
\usepackage{amsfonts}
\usepackage[utf8]{inputenc}
\usepackage{multirow} 
\usepackage{multirow}
\usepackage[table,xcdraw]{xcolor}
\usepackage{colortbl}
\usepackage{pifont}
\usepackage{bbding}

\DeclareCaptionStyle{ruled}{labelfont=normalfont,labelsep=colon,strut=off} 
\floatstyle{ruled}
\newfloat{listing}{tb}{lst}{}
\floatname{listing}{Listing}
\title{CEIDM: A Controlled Entity and Interaction Diffusion Model for Enhanced Text-to-Image Generation}

\author{
    Mingyue Yang\textsuperscript{\rm 1},
    Dianxi Shi\textsuperscript{\rm 2}\thanks{Corresponding author.},
    Jialu Zhou\textsuperscript{\rm 1},
    Xinyu Wei\textsuperscript{\rm 1},
    Leqian Li\textsuperscript{\rm 1},
    Shaowu Yang\textsuperscript{\rm 1},
    Chunping Qiu\textsuperscript{\rm 3}  
}
\affiliations{
    \textsuperscript{\rm 1}College of Computer Science and Technology, National University of Defense Technology, Changsha, China \\
    \textsuperscript{\rm 2}Advanced Institute of Big Data, Beijing, China \\
    \textsuperscript{\rm 3}Intelligent Game and Decision Lab, Beijing, China \\

    \{yangmingyue216, dxshi, zhoujialu23, xinyuwei, llq, shaowu.yang\}@nudt.edu.cn, \\
    chunping.qiu@aliyun.com

}

\usepackage{bibentry}

\begin{document}

\maketitle

\begin{abstract}
In Text-to-Image (T2I) generation, the complexity of entities and their intricate interactions pose a significant challenge for T2I method based on diffusion model: how to effectively control entity and their interactions to produce high-quality images. To address this, we propose CEIDM, a image generation method based on diffusion model with dual controls for entity and interaction. First, we propose an entity interactive relationships mining approach based on Large Language Models (LLMs), extracting reasonable and rich implicit interactive relationships through chain of thought to guide diffusion models to generate high-quality images that are closer to realistic logic and have more reasonable interactive relationships. Furthermore, We propose an interactive action clustering and offset method to cluster and offset the interactive action features contained in each text prompts. By constructing global and local bidirectional offsets, we enhance semantic understanding and detail supplementation of original actions, making the model's understanding of the concept of interactive ``actions" more accurate and generating images with more accurate interactive actions. Finally, we design an entity control network which generates masks with entity semantic guidance, then leveraging multi-scale convolutional network to enhance entity feature and dynamic network to fuse feature. It effectively controls entities and significantly improves image quality. Experiments show that the proposed CEIDM method is better than the most representative existing methods in both entity control and their interaction control.
\end{abstract}

\begin{figure*}[htbp]      
	\centering
    \includegraphics[width=1.0\textwidth]{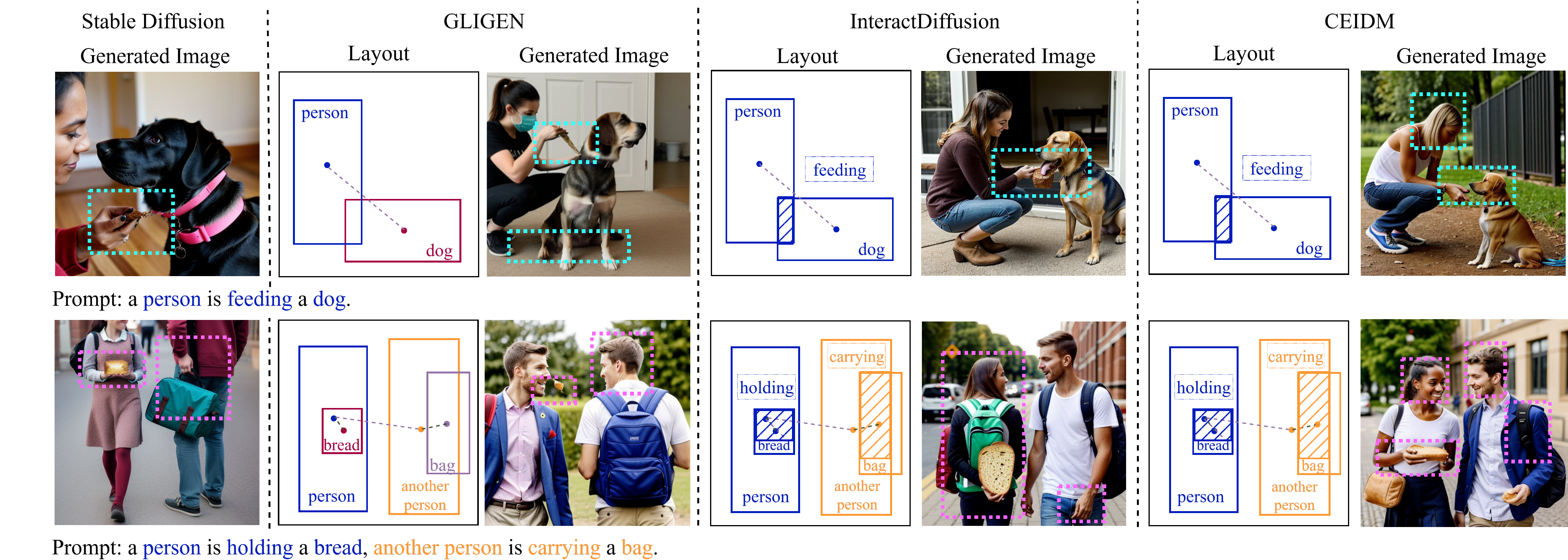}  
	\caption{Generated samples. SD input conditions only include prompt, GLIGEN adds layout, InteractDiffusion incorporates triple interactive information, while our proposed CEIDM integrates more interactive information. The shaded area in the layout represents interaction.}   
	\label{1}
\end{figure*}


\section{Introduction}

In the field of Artificial Intelligence (AI), \textbf{Text-to-Image (T2I)} technology has made rapid advancements. This technology empowers models to transform text descriptions into visual images, opening up new creative pathways for various industries. Diffusion models as the mainstream T2I models, can reconstruct the original data distribution \cite{yao2025reconstruction} and generate diverse high-quality images\cite{sohl2015deep, ho2020denoising, song2020denoising, rombach2022high, esser2021taming}, but controlling the generated content is crucial. Numerous studies have explored methods to regulate image generation in diffusion models through class\cite{dhariwal2021diffusion,zheng2022entropy}, text \cite{nichol2021glide, radford2021learning, rombach2022high, ramesh2022hierarchical,saharia2022photorealistic, zhao2023null}, image(including edge, line, scribble, and skeleton)\cite{zhang2023adding,bansal2023universal,huang2023composer}, and layout\cite{li2023gligen,zheng2023layoutdiffusion,bansal2023universal,yang2023reco, chen2024training}. However, these methods are often inadequate to generate high-quality images that meet the requirements of practical applications when faced with scenarios involving multiple entities and their complex interaction control. 

In recent years, to address the issue of insufficient interaction control, the researchers proposed a series of methods and achieved remarkable results. Among them, InteractGAN\cite{gao2020interactgan} leverages human poses and reference images for generation. Layout-based approaches\cite{hua2021exploiting} synthesize scenes layout suggestions via interactive triplets. GLIGEN\cite{li2023gligen} employs layout conditioning. InteractDiffusion\cite{hoe2024interactdiffusion} leverages interactive triplets and bounding boxes. Interfusion\cite{dai2024interfusion} extracts 3D poses as anchors for interactive scene generation. On the other hand, addressing the problem of insufficient entity control, MIGC\cite{zhou2024migc} decomposes multi-entity tasks  to control each entity. AeroGen\cite{tang2025aerogen} utilizes semantic embeddings for entity layout guidance. 

Generally, although the above methods can control entities or interactions to some extent, it is still difficult to ensure the rationality of generated interactive relationships\cite{hoe2024interactdiffusion,wang2025designdiffusion} and the accuracy of actions, and cannot effectively control entities and their interactions at the same time. As shown in \textbf{Fig.\ref{1}}, When generating the scene of ``a person is feeding a dog", there may be unreasonable interactive relationships such as ``person" or ``the dog" is suspended, ``person" has no ``food" in his hands, and ``food" is pointed to incorrectly. Even if the generated interactive relationship is reasonable, the accuracy of the generated interactive action is difficult to meet the requirements. For example, the ``feeding" action is generated as ``grabbing food" or ``throwing food", and the ``carrying" action is generated as ``pushing" or ``pulling". 
Moreover, these methods also perform poorly in controlling the generation of individual entities\cite{li2023gligen}. For example, ``person" generates six fingers or three legs, ``dog" generates two tails, and ``person" face and hands are blurred or distorted. Therefore, how to simultaneously control the generation of entities themselves and their interactions is still a challenging problem.

To address these challenges, we proposes a image generation method called \textbf{Controlled Entity and Interaction Diffusion Model (CEIDM)} that simultaneously controls entities and their interactions, aiming to generate high-quality images with a high degree of rationality. First, we leverage the Large Language Models (LLMs) by constructing chain of thought to excavate reasonable implicit interactive relationships within text. It serves as an additional guidance condition for the models, which makes the generated image content closer to the real logic and effectively improves the rationality of the generated interactive relationships.

Secondly, we enhanced and refined interactive actions through multi-scale perturbations of action features based on clustering. It as model's control conditions, enabling model to learn multi-level semantic relationships and improving its ability to distinguish complex interactions. This approach significantly enhances the model's accuracy in understanding ``action" concepts within prompt contexts, so that the fault tolerance rate of generating interactive action is higher.

Finally, on the basis of controlling the interaction, we designed an entity control network to achieve fine-grained control generation of the entity itself. Specifically, we propose a semantically guided soft attention mask generation strategy for entities, which leverage semantic information to locate multiple entities on each visual feature map and generate masks. Then we dynamically fuses the feature maps of each entity mask enhanced by multi-scale convolutional network to strengthen the model's ability to pay attention to the entity region and improve the quality of entity generation. Experimental results show that our method surpasses the existing baseline methods in terms of evaluation indicators.

Our main contributions are summarized as follows:
\begin{itemize}
\item We propose an entity implicit interactive relationship mining method based on LLMs. By taking implicit interactive information as the guiding condition of diffusion model, we can generate high-quality images with more reasonable interactive relationships.  
\item While the interactive relationship is reasonable, we further propose a clustering and offset approach for interactive actions. By constructing global-local bidirectional offset, we can strengthen the semantics and supplement the details of original actions, so that the model can generate images with more accurate interactive actions. 
\item We design an entity control network. It aims to guide the mask generation through entity semantics, and integrate the feature-enhanced multi-scale convolutional network and feature dynamic fusion network to realize the effective control of each entity, so as to further improve the quality of image generation. 
\end{itemize}
\begin{figure*}      
	\centering
    \includegraphics[width=1.0\textwidth]{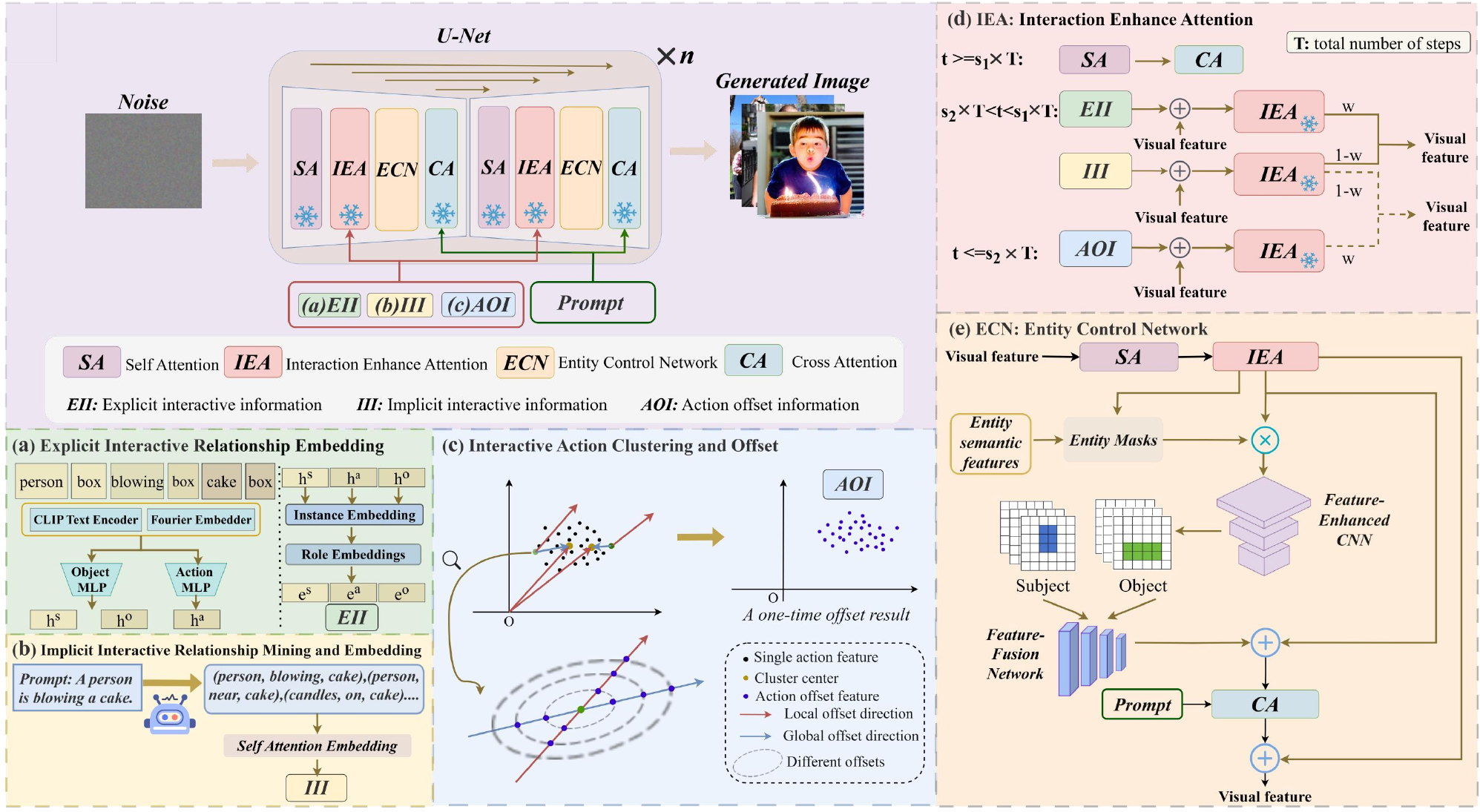}  
	\caption{The overall framework of our approach. Our method consists of (a) Explicit Interactive Relationship Embedding, (b) Implicit Interactive Relationship Mining and Embedding, (c) Interactive Action Clustering and Offset, (d) Interaction Enhance Attention (IEA), and (e) Entity Control Network (ECN).} 
	\label{2}
\end{figure*}

\section{Method}

The diffusion image generation method \textbf{CEIDM} proposed in this paper aims to solve the problems of insufficient control over entities and interactions in T2I, its framework is shown in \textbf{Fig.\ref{2}}. It is mainly composed of the following parts: (a) Explicit interactive relationship embedding, which is mainly used to process the triple described in the text prompt to obtain explicit interactive information. (b) Implicit interactive relationship mining mechanism based on LLMs that mines the reasonable and rich implicit interactive relationships contained in the text prompt, and realizes embedding through self-attention to obtain the implicit interactive information. (c) Interactive action clustering and offset mechanism that enhances action categories and supplement action details to obtain action offset interactive information. (d) Interaction attention layer that is used to process various interactive information. (e) Entity control network that enables fine-grained entity control. In summary, we introduce explicit interactive, implicit interactive information and action offset information into the diffusion model through the interaction attention layer to strengthen the control of interaction. Meanwhile, the entity control network is used to realize the control of entity generation. Our Interaction Enhance Attention (IEA) and Entity Control Network (ECN) can be seamlessly integrated into existing T2I diffusion models(Diffusion Model's Basic Knowledge see \textbf{Appendix B}).

\subsection{Explicit Interactive Relationship Embedding}

As show in \textbf{Fig.\ref{2}(a)}, we encode\cite{devlin2019bert, mildenhall2021nerf, hoe2024interactdiffusion} explicit interactive relationships in text prompt, which are already expressed as triples in the HICO-DET\cite{chao2018learning} dataset: 
\begin{equation}  
h_i^s,h_i^o=MLP_{Obj}([CLIP(phrases),F(boxes)])
\label{e1}
\end{equation}
\begin{equation}  
h_i^a=MLP_{Act}([CLIP(phrases),F(boxes)])
\label{e2}
\end{equation}
\begin{equation}  
(e_i^s,e_i^a,e_i^o)=(h_i^s+q_i+r^s,h_i^a+q_i+r^a,h_i^o+q_i+r^o)
\label{e3}
\end{equation}

Where $i$ represents the number of interaction instances in a text prompt. ${CLIP}(\cdot)$ denotes semantic embedding of $subject\_phrases$,$action\_phrases$,$object\_phrases$. ${F}(\cdot)$ indicates Fourier Embedding of corresponding bounding boxes. $q_i$ stands for instance embedding, where all tokens within an interaction triplet instance share the same instance embedding. ${r^{s},r^{a},r^{o}}$ respectively denote role embeddings for subject, action, and object. $e_i^s,e_i^a,e_i^o$ represent the final embedding results of subject, action and object respectively, namely explicit interactive information. 

\subsection{Implicit Interactive Relationship Mining and Deep Semantic Embedding}

To enhance the diffusion model's understanding of interaction scenarios and ensure more logical relationships in generated images, we strengthened conditional descriptions of interactive relationships through fully leveraging the profound text comprehension and powerful reasoning capabilities of LLMs\cite{tang2024fusing, ding2025enhancing}. This approach provides richer and more reasonable implicit interactive relationships to guide the denoising process. Specifically, as shown in \textbf{Fig.\ref{2}(b)}, we conducted in-depth mining of interactive relationships in text prompt based on chain of though, returning them as triplet (It can be incorporated into the HICO-DET to form a \textbf{New Dataset}). The prompt templates are detailed in \textbf{Appendix C}. The LLMs' output mainly contains some common rational prior knowledge implied in the text prompt. For example, when the prompt is ``A person is blowing a cake", the LLMs can extracts obvious interactive relationships (person, blowing, cake), and uncovers multiple implicit ones: the person should be near the cake (person, near, cake); and the person's mouth should face the flame (person's mouth, directed at, flame), etc. 
\begin{figure}[t!]      
	\centering
    \includegraphics[scale=0.2]{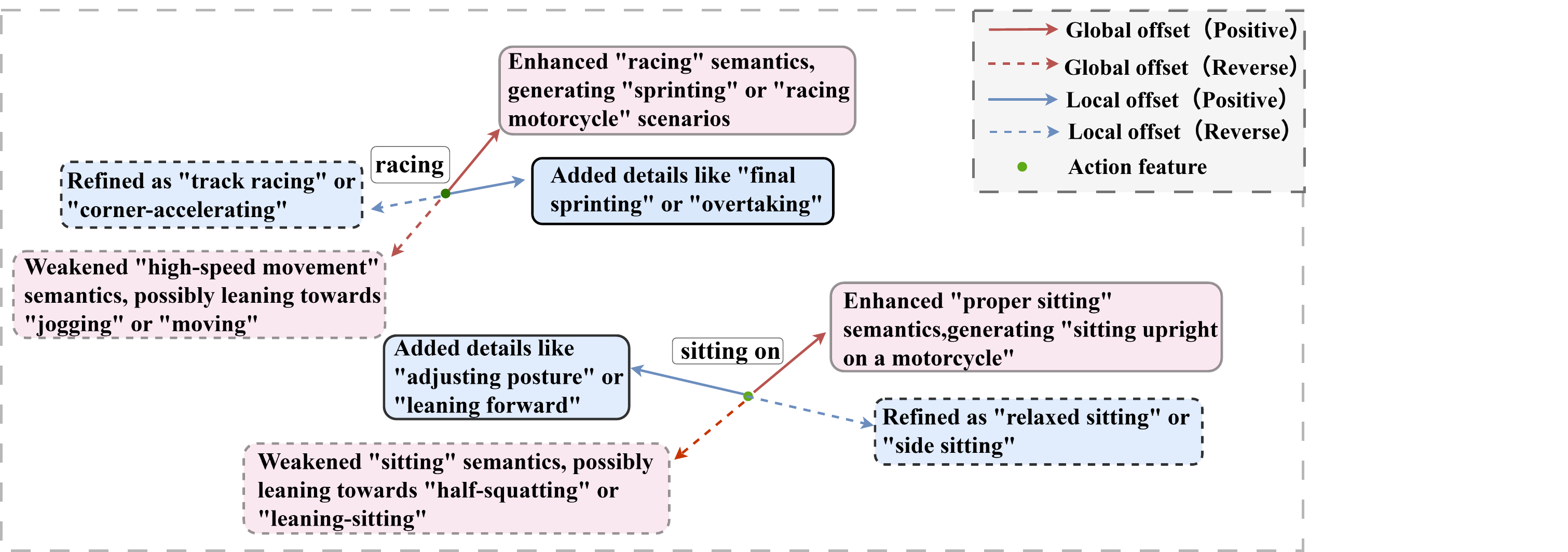}  
	\caption{Partial decoding results after interactive action feature offset. In this case, the reverse offset introduced in each direction may reduce the influence of some features.}
	\label{3}
\end{figure}

We perform deep embedding on the generated triplets to enhance the diffusion model's understanding of interactive relationship. For a relational triplet $(h,r,t)$, we first apply three independent linear transformation layers to model the different roles semantic (subject, action, object) separately, obtaining $(h_{emb},r_{emb},t_{emb})$. Then, we leverage self-attention to capture the internal relationship ($resi$) within the triplet. Finally, residual connections are applied to integrate $resi$ into each role's original embedding, yielding the final deep semantic embedding results $({e}^{h},{e}^{{r}},{e}^{{t}})$, which namely implicit interactive information. 
\begin{equation}  
resi=Self\_Attention(Stack(h_{emb},r_{emb},t_{emb}))
\label{e4}
\end{equation}
\begin{equation}  
(e^{h},e^{r},e^{t})=(h_{emb}+resi,r_{emb}+resi,t_{emb}+resi)
\label{e5}
\end{equation}

\subsection{Interactive Action Clustering and Offset}

While generating reasonable interactive relationships, it is also crucial to ensure the accuracy of the generated interactive actions. Most existing methods based on diffusion models struggle to comprehend the semantics of  ``action", often resulting in inaccurate action generation. This leads to discrepancies between the generated image actions and the textual action semantics, poor overall generation quality of the images, and difficulties in detecting interactive actions. As shown in \textbf{Fig.\ref{2}(c)}, this section focuses on the interactive ``action". Since the action semantics in the text prompts of HICO-DET\cite{chao2018learning} are highly overlapping, we alleviate the above problem by introducing clustering and offset of action semantics features.

Taking prompt ``a person is racing a motorcycle, a person is riding a motorcycle, a person is sitting on a motorcycle, a person is straddling a motorcycle, a person is turning a motorcycle" as an example, we first leverage CLIP to embed all interactive actions as ${A_{i}}$, where $i$ represents the number of actions. Then, we introduce the K-means algorithm to cluster all action features in ${A_{i}}$ : Dynamic Motion Class (racing, riding, straddling, turning) and Static Pose Class (sitting on). Finally, under multiple offset values, each action feature in ${A_{i}}$ is offset along two directions: global (the direction of the cluster center which it belongs) and local (the direction of the unit vector to its cluster center). 

The offset processes of the ``racing" and ``sitting on" actions, along with partial decoding results, are shown in \textbf{Fig.\ref{3}}. To further demonstrate the action offset effects of our method, we present partial decoding outcomes for the ``carrying" and ``wearing" actions in \textbf{Table \ref{t1}} and \textbf{Table \ref{t2}}. A detailed analysis is provided in \textbf{Appendix D}.
\begin{table}[t!]
\small
\centering
\begin{tabular}{c|l|l}
\hline
\textbf{Action} & \multicolumn{1}{c|}{\textbf{Positive offset (+0.1)}}                          & \multicolumn{1}{c}{\textbf{Reverse offset (-0.1)}}                           \\ \hline
carrying        & \begin{tabular}[c]{@{}l@{}}transporting, hauling, \\ moving\end{tabular}     & \begin{tabular}[c]{@{}l@{}}releasing, dropping, \\ placing\end{tabular}     \\ \hline
wearing         & \begin{tabular}[c]{@{}l@{}}dressing, accessorizing, \\ adorning\end{tabular} & \begin{tabular}[c]{@{}l@{}}removing, undressing, \\ taking off\end{tabular} \\ \hline
\end{tabular}
\caption{Example of the global offset of interactive action.}
\label{t1}
\end{table}
\begin{table}[t!]
\centering
\small
\begin{tabular}{c|l|l}
\hline
\textbf{Action} & \multicolumn{1}{c|}{\textbf{Positive offset (+0.05)}}                                  & \multicolumn{1}{c}{\textbf{Reverse offset (-0.05)}}                                       \\ \hline
carrying        & \begin{tabular}[c]{@{}l@{}}hand-carrying, shoulder\\ -carrying, toting\end{tabular}    & \begin{tabular}[c]{@{}l@{}}light-carrying, brief\\ -holding, partial-lifting\end{tabular} \\ \hline
wearing         & \begin{tabular}[c]{@{}l@{}}tight-wearing, layered\\ -wearing, strapping\end{tabular} & \begin{tabular}[c]{@{}l@{}}loose-wearing, partial\\ -covering, draping\end{tabular}       \\ \hline
\end{tabular}
\caption{Example of the local offset of interactive action.}
\label{t2}
\end{table}


As can be seen from the offset results: The global offset aligns each action feature in $A_i$ with its corresponding semantic category, thereby enhancing the commonality of basic action categories. The local offset allows fine-tuning of specific details such as posture, force intensity, or angle variations of individual actions. This bidirectional offset preserves category commonality while enriching contextual details. This method enables the model to learn richer semantic distributions of actions and avoid overfitting a single feature. 

By combining different offsets and directions, we can obtain action features with similar semantics to $A_i$: ${A_i^1,\cdots,A_i^j,\cdots,A_i^m}$. Where ${j\in[1,m],j\in N}$ and $m=offset$ $quantity*2$. Before inputting $A_i^j$ into the diffusion model, we first further embed it(\textbf{Eq.(\ref{e1})(\ref{e2})(\ref{e3})}) to obtain $e_{i}^{A^{j}}$, then combine $e_{i}^{A^{j}}$ with the corresponding original subject information ${e_i^s}$ and object information ${e_i^o}$ respectively to obtain the complete interactive information. 
\begin{equation}  
G_{0}=Cat({e_{i}^{s},e_{i}^{a},e_{i}^{o}}), G_{j}=Cat({e_{i}^{s},e_{i}^{A^{j}},e_{i}^{o}})
\label{e6}
\end{equation}

Where, $e_{i}^{a}$ represents the original action tokens corresponding to $e_{i}^{A^{j}}$ without any offset. We connect $G_0,G_1,\cdots,G_j,\cdots,G_m$ to form the action offset interactive information $G$. $G$ is used as the supplement of interactive actions to guide the diffusion generation process, which can enhance the model's ability to express subtle semantic changes of interactive actions and improve the model's adaptability and robustness to complex scenarios. 

\subsection{Interaction Enhance Attention (IEA)}

As shown in \textbf{Fig.\ref{2}(a)}, we add a Gated Self-Attention layer\cite{li2023gligen,hoe2024interactdiffusion} into the LDM's Transformer to process various interactive information obtained above, while introducing an interaction scaling coefficient $\delta$\cite{chung2024style} to enhance the model's focus on key interaction features. Let ${v}=\begin{bmatrix}\mathbf{v}_{1},\mathbf{v}_{2},\cdots\end{bmatrix}$ represent the visual feature tokens of the image during denoising. The attention mechanism is defined as: 
\begin{equation}  
StrenAtt(Q,K,V)=Softmax(\frac{Q(K)^T}{\sqrt{d}})\cdot V\cdot\delta
\label{e7}
\end{equation}

Where $Q=W_Q\cdot[v,info]$,$K=W_K\cdot[v,info]$,$V=W_V\cdot[v,info]$.$[\cdot,\cdot]$ indicates that all tokens are connected. $info$ can represent three types of embedded information: (a) Explicit interactive information $({e_{i}^{s},e_{i}^{a},e_{i}^{o}})$; (b) Implicit interactive information$({e_{i}^{h},e_{i}^{r},e_{i}^{t}})$; (c) Action offset interactive information $G$.
\begin{figure}      
	\centering
        \includegraphics[width=1.0\linewidth]{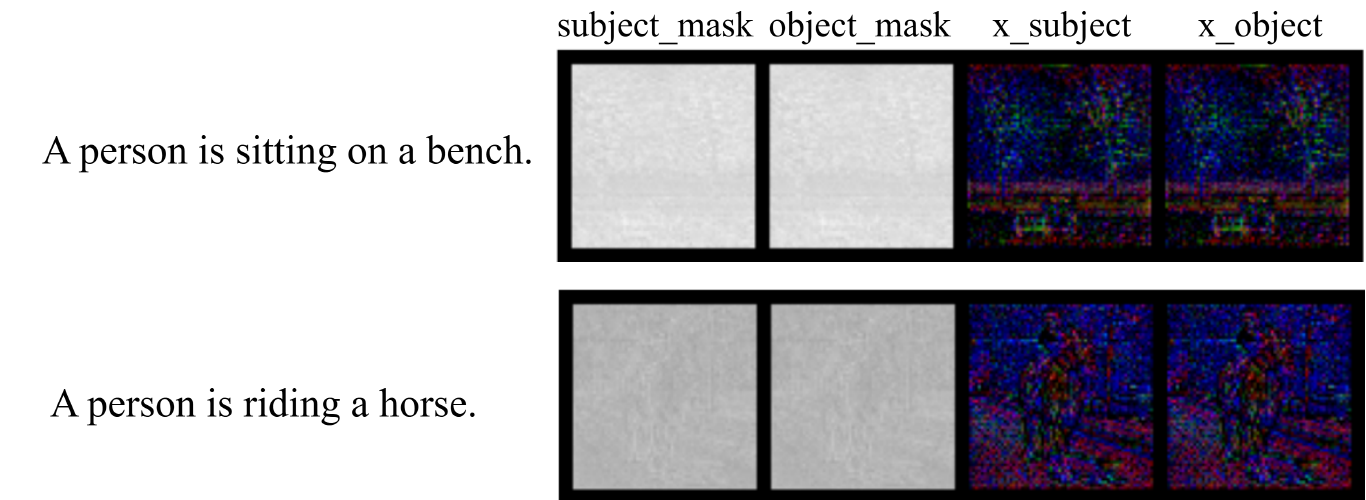}  
	\caption{Visualization of subject and object's soft attention masks and their mask results. }
	\label{4}
\end{figure}

We design the sampling interval control strategy shown in \textbf{Fig.\ref{2}(d) }(introduce hyperparameters $s_1$ and $s_2$). Fusion weights $w$ is introduced to process the output of the IEA. The design reasons and details are shown in \textbf{Appendix E}. 

\subsection{Entity Control Network (ECN)}

In controlling the generation of interactions, we also controlled the generation of each entity by adding an entity control network after IEA layer. As shown in \textbf{Fig.\ref{2}(e)}, this work mainly consists of the following three parts. 

\textbf{Entity Semantic-Driven Dynamic Localization. }Our primary focus is on the regions containing subject and object entities. First, we employ a Multilayer Perceptron (MLP) to process CLIP semantic embeddings of entities. The MLP outputs are then combined with image visual features through fusion modeling to generate soft attention masks ($subject\_mask$,$object\_mask$). Subsequently, the visual features from the IEA layer are masked to obtain subject and object mask features ($x\_subject$, $x\_object$). Visualization example is shown in \textbf{Fig.\ref{4}}. The soft masks enable the model to preserve key detailed features while maintaining awareness of background information. We introduce a temperature regulation coefficient $temp$ during the mask generation to control the distribution morphology of the masks.


\textbf{Entity Features-Enhanced Multi-scale Convolutional Network.} The core is to refine the subject and object regions selected by dynamic focus through parallel multi-scale convolution branches, and extract and fuse the entity feature information of different receptive fields. 

\textbf{Entity Feature Dynamic Fusion Network}. We will integrate the enhanced subject and object features. The network dynamically adjusts the weight of convolution kernel by input features, so that the model can adaptively optimize the feature fusion effect under different input, and improve the semantic coherence and detail quality of entity fusion. 

Besides, As shown in \textbf{Fig.\ref{2}(a)(e)}, we combine the features from the entity fusion network with the visual features output by the IEA layer, which as visual features input for the CA. Our method also requires that the output residual of the CA layer be directly connected to the IEA layer's output to ensure the coherence of the generated image content. 

\section{Experiments}
\subsection{Experimental Setting}
\textbf{Datasets.} HICO-DET dataset\cite{chao2018learning} comprises 47,776 images: 38,118 for training and 9,658 for testing. It includes 600 types of HOI triplets constructed from 80 object categories and 117 verb classes. In experiments, We extracted the annotations in the testing set (includes 33,405 HOI annotations) as input to generate images.
\begin{figure}     
	\centering
        \includegraphics[width=1.0\linewidth]{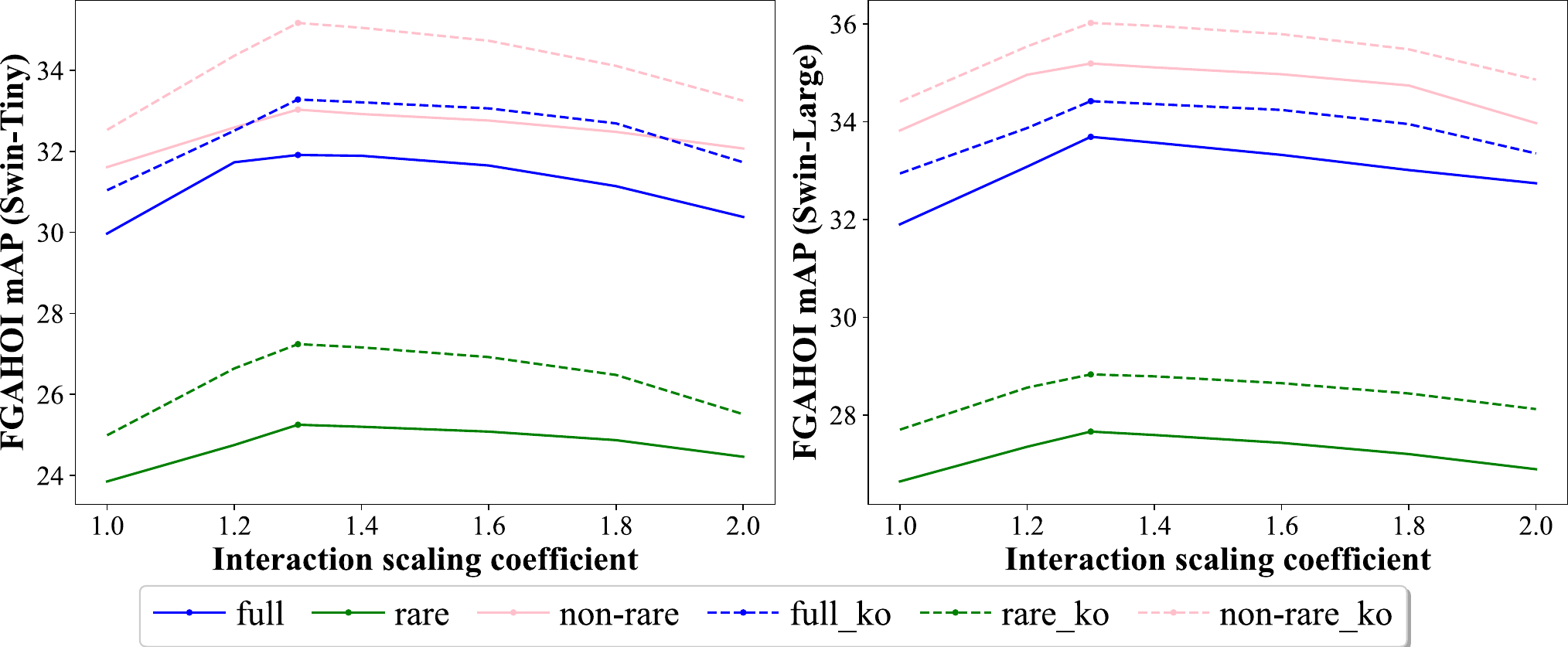}  
	\caption{The left and right figures respectively show the HOI detection scores under different $\delta$ measured using FGAHOI with Swin-Tiny and Swin-Large backbone.}
	\label{5}
\end{figure}
\begin{figure}  
	\centering
    \includegraphics[width=1.0\linewidth]{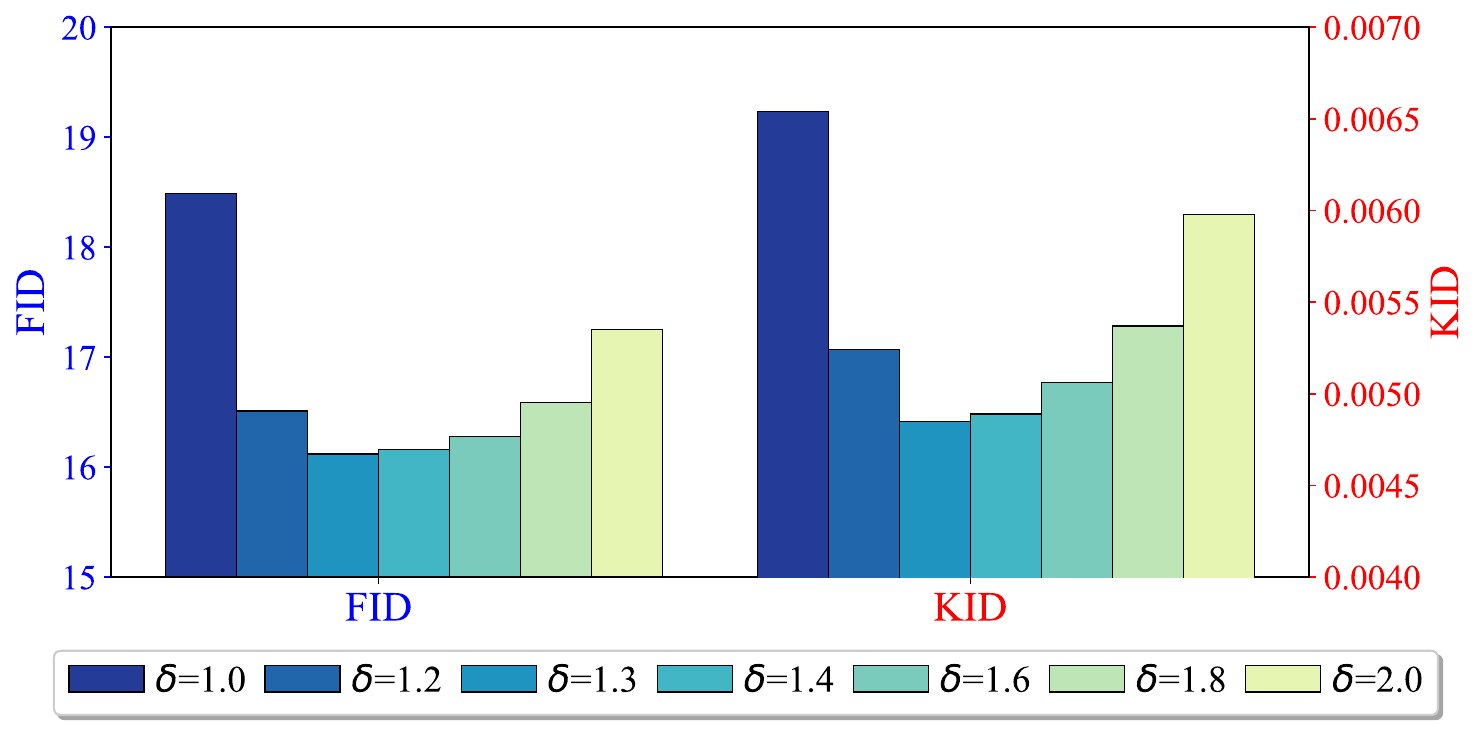}  
	\caption{Quality scores under different $\delta$.}
	\label{6}
\end{figure} 

\begin{figure*}[t!]      
	\centering
  \includegraphics[width=0.94\linewidth]{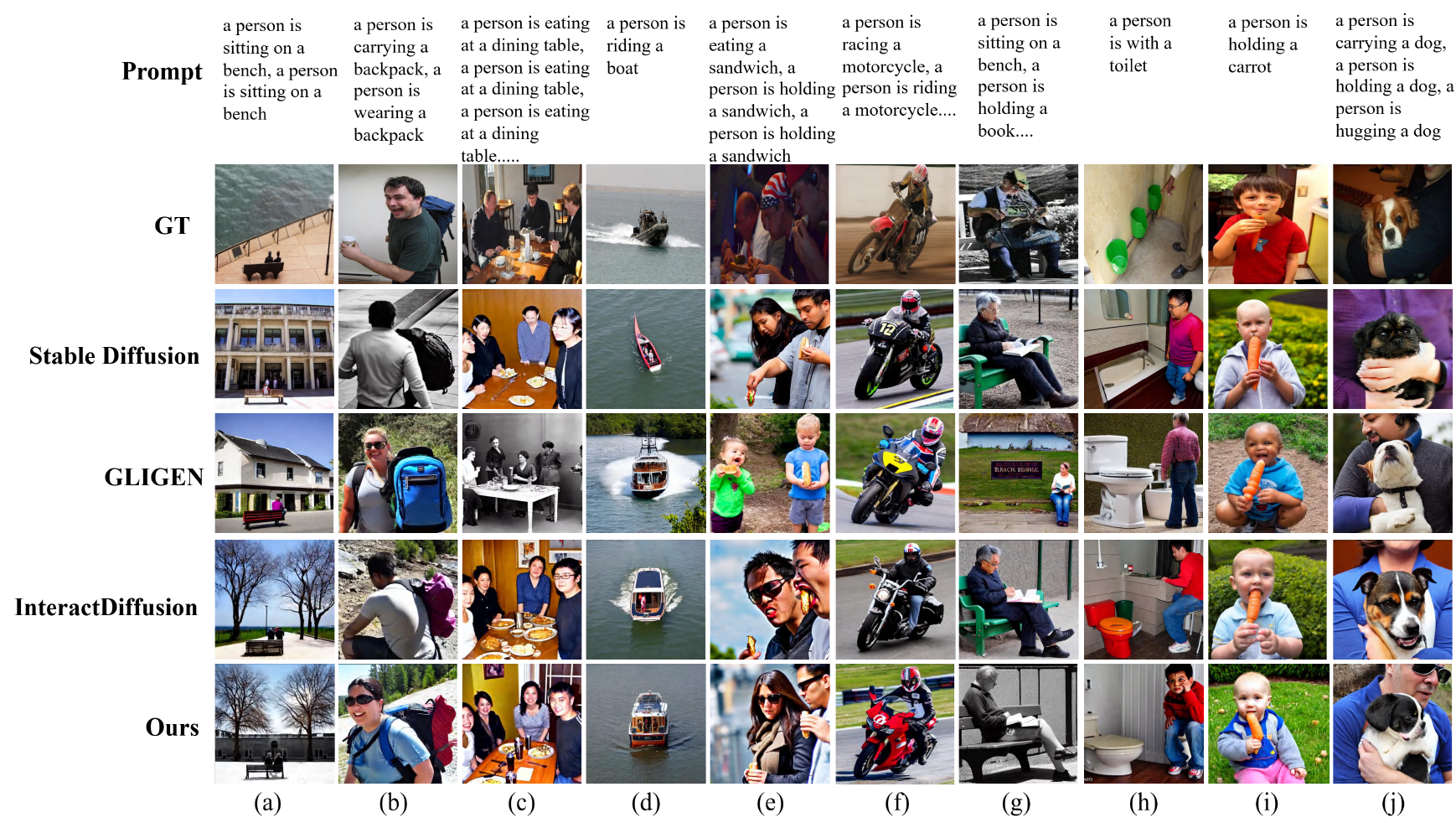}  
	\caption{Our method is qualitatively compared with existing baselines. GT represents the ground truth images.}
	\label{7}
\end{figure*}
\begin{table*}[t!]
\scriptsize
\small
\centering
\setlength{\tabcolsep}{7.3pt} 
\begin{tabular}{l|cc|cccc|cccc}
\hline
\multicolumn{1}{c|}{\multirow{3}{*}{\textbf{Model}}} & \multicolumn{2}{c|}{\multirow{2}{*}{\textbf{Quality ↓}}} & \multicolumn{4}{c|}{\textbf{Swin-Tiny(mAP)↑}}                                            & \multicolumn{4}{c}{\textbf{Swin-Large(mAP)↑}}                                            \\ \cline{4-11} 
\multicolumn{1}{c|}{}                                & \multicolumn{2}{c|}{}                                    & \multicolumn{2}{c|}{Default}                         & \multicolumn{2}{c|}{Known Object} & \multicolumn{2}{c|}{Default}                         & \multicolumn{2}{c}{Known Object} \\ \cline{2-11} 
\multicolumn{1}{c|}{}                                & FID                        & KID                         & Full           & \multicolumn{1}{c|}{Rare}           & Full            & Rare            & Full           & \multicolumn{1}{c|}{Rare}           & Full            & Rare           \\ \hline
StableDiffusion                                      & 35.85                      & 0.01297                     & 0.63           & \multicolumn{1}{c|}{0.68}           & 0.66            & 0.70            & 0.64           & \multicolumn{1}{c|}{0.83}           & 0.65            & 0.84           \\
GLIGEN                                               & 29.35                      & 0.01275                     & 21.73          & \multicolumn{1}{c|}{15.35}          & 23.31           & 17.24           & 23.99          & \multicolumn{1}{c|}{19.56}          & 24.99           & 20.37          \\
InteractDiffusion                                    & \underline{18.69}                      & \underline{0.00676}                     & \underline{29.53}          & \multicolumn{1}{c|}{\underline{23.02}}          & \underline{30.99}           & \underline{24.93}           & \underline{31.56}          & \multicolumn{1}{c|}{\underline{26.09}}          & \underline{32.52}           & \underline{27.04}          \\ \hline
\rowcolor{gray!20} 
\textbf{Ours}                                        & \textbf{{16.12}}             & \textbf{{0.00485}}            & \textbf{{31.91}} & \multicolumn{1}{c|}{\textbf{{25.27}}} & \textbf{{33.28}}  & \textbf{{27.24}}  & \textbf{{33.69}} & \multicolumn{1}{c|}{\textbf{{27.66}}} & \textbf{{34.61}}  & \textbf{{28.83}} \\ \hline  \hline
HICO-DET GT                                          & /                          & /                           & 29.94          & \multicolumn{1}{c|}{22.24}          & 32.48           & 24.16           & 37.18          & \multicolumn{1}{c|}{30.71}          & 38.93           & 31.93          \\ \hline
\end{tabular}
\caption{Quantitative comparison. We use bold and underlines to represent the best and second-best results, respectively.}
\label{t3}
\end{table*}

\textbf{Evaluation Metrics.} We employ the Frechet Inception Distance(\textbf{FID})\cite{heusel2017gans} and Kernel Inception Distance(\textbf{KID})\cite{binkowski2018demystifying} to evaluate both the overall quality of image generation and the controllability of entities and interactions in the generated models. When the quality of all entities and interactions improves, the overall image quality will correspondingly enhance. The \textbf{HOI Detection Score} is used to quantify the models’ controllability in interaction generation(interactive relationship's rationality and action's accuracy). To assess this, we utilize the pre-trained HOI detector FGAHOI\cite{ma2023fgahoi} (Swin-Tiny and Swin-Large backbone) to detect HOI instances within the generated images. See \textbf{Appendix F} for details.

\textbf{Implementation Details.} Our method is free of training. All experiments use the InteractDiffusion\cite{hoe2024interactdiffusion} generation model. We leveraged PLMS\cite{liu2022pseudo} sampler with 50 sampling steps. During evaluation, we generated images using a NVIDIA A100-SXM4-40GB GPU, with inference time about 185 hours. Our random seed (default 489) controls noise generation, data preprocessing and order shuffling. Extensive comparative experiments revealed that LLMs employing Qwen-Turbo\cite{yang2025qwen3} effectively extracted latent textual relationships. Additionally, we conducted sensitive analysis by adjusting the interaction scaling coefficient $\delta$ from 1.0 to 2.0 in 0.1 increments (see \textbf{Fig.\ref{5}} and \textbf{Fig.\ref{6}}). More hyperparameter sensitivity analysis shown in \textbf{Appendix G}. We implemented: (1) the IEA layer with $\delta=1.3$ and feature output fusion weight $w=0.7$; (2) sampling strategy hyperparameters $s_{1}=1,s_{2}=0.7$; (3) $temp=2.0$ in attention mask computation. 

\subsection{Qualitative Analysis}

\textbf{Fig.\ref{7}} shows a qualitative comparison. We use the prompt format of ``a person $<$action$>$ a $<$object$>$". Experimental results show that there are many problems in the images generated by other models, while our model shows significant advantages in presenting the entity and its interactions.

For instance, in \textbf{Fig.\ref{7}}, Stable Diffusion (SD) and InteractDiffusion generated incorrect interactive action (``sitting on") in column (a). SD and InteractDiffusion generated unreasonable interactive relationship (``carrying") in column (b). The ``backpack" is even suspended. In column (j), the hand details in images generated by SD, GLIGEN, and InteractDiffusion were severely distorted or partially obscured. The interaction details in the images generated by other methods in columns (c)-(i) are not accurate enough. Our proposed method effectively mitigates these issues, delivering more accurate representation of interaction details. When examining multiple interactions, as shown in column (c) of \textbf{Fig.\ref{7}}, only our method can reasonably present all the interactive relationships and generate more accurate interactive actions. Furthermore, columns (c)(e)(g)(h)(i) of \textbf{Fig.\ref{7}} reveal low quality in other methods 'generated entities: blurred entities and distorted facial features in column (c); the person entity generated by InteractDiffusion has three legs in column (g). But our method realizes the constraint control of entities and improves the generation quality of each entity. 
\begin{table*}[t!]
\small
\centering
\setlength{\tabcolsep}{3.8pt}
\begin{tabular}{l|lllllll|cc|cccc|cccc}
\hline
\multicolumn{1}{c|}{\multirow{3}{*}{\textbf{Model}}} & \multicolumn{1}{c}{\multirow{3}{*}{\textbf{Tr.}}} & \multicolumn{1}{c}{\multirow{3}{*}{\textbf{To.}}} & \multicolumn{1}{c}{\multirow{3}{*}{\textbf{Em.}}} & \multicolumn{1}{c}{\multirow{3}{*}{\textbf{Cot.}}} & \multicolumn{1}{c}{\multirow{3}{*}{\textbf{Cl.}}} & \multicolumn{1}{c}{\multirow{3}{*}{\textbf{Ec.}}} & \multicolumn{1}{c|}{\multirow{3}{*}{\textbf{$\delta.$}}} & \multicolumn{2}{c|}{\multirow{2}{*}{\textbf{Quality ↓}}} & \multicolumn{4}{c|}{\textbf{Swin-Tiny (mAP)  ↑}}                                         & \multicolumn{4}{c}{\textbf{Swin-Large (mAP)  ↑}}                                        \\ \cline{11-18} 
\multicolumn{1}{c|}{}                                & \multicolumn{1}{c}{}                              & \multicolumn{1}{c}{}                              & \multicolumn{1}{c}{}                              & \multicolumn{1}{c}{}                               & \multicolumn{1}{c}{}                              & \multicolumn{1}{c}{}                              & \multicolumn{1}{c|}{}                             & \multicolumn{2}{c|}{}                                    & \multicolumn{2}{c|}{Default}                         & \multicolumn{2}{c|}{Known Object} & \multicolumn{2}{c|}{Default}                         & \multicolumn{2}{c}{Known Object} \\ \cline{9-18} 
\multicolumn{1}{c|}{}                                & \multicolumn{1}{c}{}                              & \multicolumn{1}{c}{}                              & \multicolumn{1}{c}{}                              & \multicolumn{1}{c}{}                               & \multicolumn{1}{c}{}                              & \multicolumn{1}{c}{}                              & \multicolumn{1}{c|}{}                             & FID                        & KID                         & Full           & \multicolumn{1}{c|}{Rare}           & Full            & Rare            & Full           & \multicolumn{1}{c|}{Rare}           & Full            & Rare           \\ \hline
StableDiffusion                                      &                                                   &                                                   &                                                   &                                                    &                                                   &                                                   &                                                   & 35.85                      & 0.01297                     & 0.63           & \multicolumn{1}{c|}{0.68}           & 0.66            & 0.70             & 0.64           & \multicolumn{1}{c|}{0.83}           & 0.65            & 0.84           \\
GLIGEN                                               & \ding{52}*                                                &                                                   &                                                   &                                                    &                                                   &                                                   &                                                   & 29.35                      & 0.01275                     & 21.73          & \multicolumn{1}{c|}{15.35}          & 23.31           & 17.24           & 23.99          & \multicolumn{1}{c|}{19.56}          & 24.99           & 20.37          \\
InteractDiffusion                                    & \ding{52}                                                 & \ding{52}                                                 & \ding{52}                                                 &                                                    &                                                   &                                                   &                                                   & 18.69                      & 0.00676                     & 29.53          & \multicolumn{1}{c|}{23.02}          & 30.99           & 24.93           & 31.56          & \multicolumn{1}{c|}{26.09}          & 32.52           & 27.04          \\ \hline
\multirow{5}{*}{\textbf{Ours}}                       & \ding{52}                                                 & \ding{52}                                                 & \ding{52}                                                 &                                                    &                                                   &                                                   &                                                   & 18.69                      & 0.00676                     & 29.53          & \multicolumn{1}{c|}{23.02}          & 30.99           & 24.93           & 31.56          & \multicolumn{1}{c|}{26.09}          & 32.52           & 27.04          \\
                                                     & \ding{52}                                                 & \ding{52}                                                 & \ding{52}                                                 & \ding{52}                                                  &                                                   &                                                   &                                                   & 18.09                      & 0.00583                     & 29.97          & \multicolumn{1}{c|}{23.56}          & 31.34           & 25.29           & 31.90           & \multicolumn{1}{c|}{26.36}          & 32.94           & 27.47          \\
                                                     & \ding{52}                                                 & \ding{52}                                                 & \ding{52}                                                 & \ding{52}                                                  & \ding{52}                                                 &                                                   &                                                   & 17.85                      & 0.00567                     & 31.18          & \multicolumn{1}{c|}{24.63}          & 32.52           & 26.68           & 32.94          & \multicolumn{1}{c|}{27.37}          & 33.99           & 28.25          \\
                                                     & \ding{52}                                                 & \ding{52}                                                 & \ding{52}                                                 & \ding{52}                                                  & \ding{52}                                                 & \ding{52}                                                 &                                                   & \underline{16.25}                      & \underline{0.00499}                     & \underline{31.64}          & \multicolumn{1}{c|}{\underline{{25.92}}}          & \underline{32.97}           & \underline{27.01}           & \underline{33.35}          & \multicolumn{1}{c|}{\underline{{27.58}}}          & \underline{34.36}           & \underline{28.69}          \\
                                                     & \ding{52}                                                 & \ding{52}                                                 & \ding{52}                                                 & \ding{52}                                                  & \ding{52}                                                 & \ding{52}                                                 & \ding{52}                                                 & \textbf{\cellcolor{gray!20}{16.12}}             & \textbf{\cellcolor{gray!20}{0.00485}}            & \textbf{\cellcolor{gray!20}{31.91}} & \multicolumn{1}{c|}{\textbf{\cellcolor{gray!20}{25.27}}} & \textbf{\cellcolor{gray!20}{33.28}}  & \textbf{\cellcolor{gray!20}{27.24}}  & \textbf{\cellcolor{gray!20}{33.69}} & \multicolumn{1}{c|}{\textbf{\cellcolor{gray!20}{27.66}}} & \textbf{\cellcolor{gray!20}{34.61}}  & \textbf{\cellcolor{gray!20}{28.83}} \\ \hline \hline
HICO-DET GT                                          &                                                   &                                                   &                                                   &                                                    &                                                   &                                                   &                                                   & /                          & /                           & 29.94          & \multicolumn{1}{c|}{22.24}          & 32.48           & 24.16           & 37.18          & \multicolumn{1}{c|}{30.71}          & 38.93           & 31.93          \\ \hline
\end{tabular}
\caption{The ablation results. $Tr.$ denotes Interaction Transformer. $To.$ represents Interaction Tokenizer. $Em.$ indicates Interaction Embedding. $Cot.$ refers to implicit interactive relationship inference and deep semantic embedding. $Cl.$ signifies clustering and offset of interactive actions. $Ec.$ denotes entity control network. \ding{52}* indicates Gated Self-Attention in GLIGEN\cite{li2023gligen, hoe2024interactdiffusion}. We use bold and underlines to represent the best and second-best results, respectively.}
\label{t4}
\end{table*}
\begin{figure*}[t!]      
	\centering
        \includegraphics[width=0.88\textwidth]{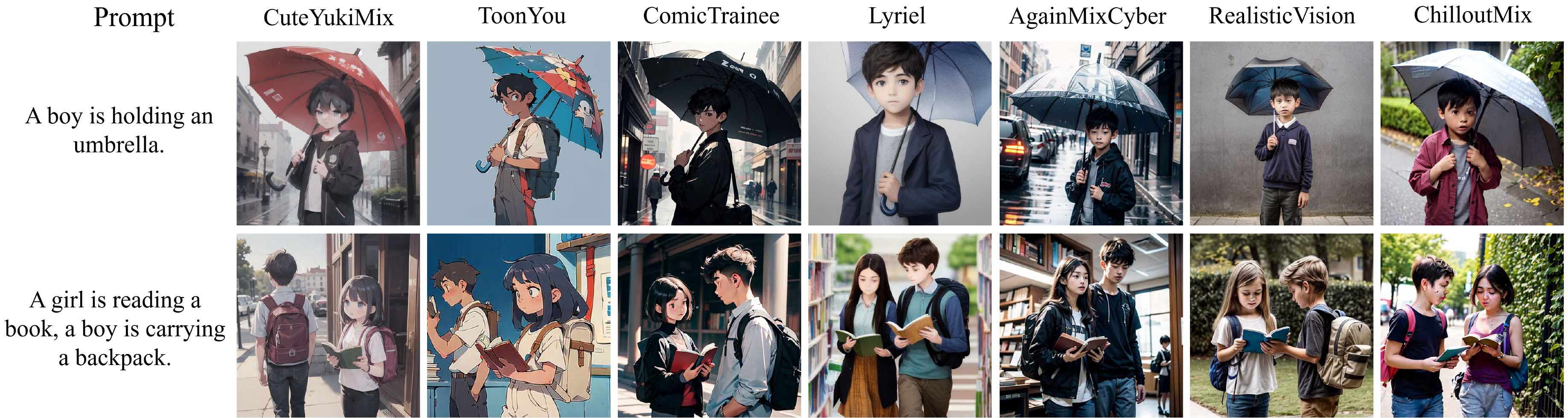}  
	\caption{Visualization of CEIDM on various personalized Stable Diffusion Models. Transferable interaction and entity control successfully retain the unique style attributes of each model, while improving the controllability of entities and interactions.}
	\label{8}
\end{figure*}

\subsection{Quantitative Analysis}

The quantitative results are shown in \textbf{Table \ref{t3}}. Compared with the existing baseline, our method enhances interaction controllability while maintaining high quality image generation capability, so it outperforms the existing methods in all indicators and obtains a certain model performance gain.

For \textbf{Image Quality}, our method demonstrates significantly superior results compared to baselines. This indicates that by incorporating extensive interactive control information into the original diffusion model, introducing an entity control network, the quality of each entity and its interaction in the generated image is greatly improved, thus improving the overall quality of the image.

For \textbf{HOI Detection Scores}, SD lacks positional priors, GLIGEN only constrains layout, and InteractDiffusion provides weak interaction guidance, all three performed poorly in the HOI benchmark. Our method greatly enriches the model's understanding of interactive scenes by introducing implicit interactive information and action offset interactive information, markedly improving HOI detection accuracy.

As shown in \textbf{Table \ref{t3}}, the improvement in the Swin-Large backbone detection score was slightly higher than that of the Swin-Tiny backbone. This shows that our method outperforms in generating complex scene images while excelling at handling image details. Additionally, when we use Swin-Tiny detection, the mAP between the generated and the real image is close, while the gap is widened under Swin-Large detection. It indicates that there is still room for further improvement in rendering finer action details in our method.

\subsection{Ablation Studies}
The ablation results in \textbf{Table \ref{t4}} show: GLIGEN's integration of Gated Self-Attention layers into the Transformer Module of the SD to incorporate additional layout conditions, which resulted in a significant reduction of FID from 35.85 to 29.35, while boosting all mAP scores by 25-40 times. The InteractDiffusion additionally introduced the interactive triple information to realize the interaction control, which reduced the FID from 29.35 to 18.69, and improved all mAP scores by 7-8 points. Building on this foundation, our method progressively added Cot., Cl., and Ec. components, reducing FID from 18.69 to 16.25 and HOI detection scores from 29.53 to 31.64. Finally, the introduction of interaction scaling coefficients $\delta$ lowered FID to 16.12, with detection scores further improved to 31.91. 

Our entity and interaction control module enable precise control over the generation process. \textbf{Fig.\ref{8}} demonstrates visualizations of our method across several personalized Stable Diffusion Models, further validating the potential of introducing entity and interaction controls without compromising the model's unique qualities. 

\section{Conclusion}
This paper proposes an innovative diffusion image generation method designed to address the limitations in entity and interaction control, namely CEIDM. By integrating explicit interactive, implicit interactive and action offset information into the diffusion model's generation process, the method significantly enhances interaction control. The newly developed ECN enables precise control of entities. Experiments demonstrate that this method outperforms existing approaches in both image overall quality and interactive detection accuracy, exhibiting excellent performance in complex scene image synthesis and fine detail processing. It provides a new idea to promote the development of T2I generation.

\bibliography{aaai2026}

\end{document}